\pdfoutput=1

\documentclass[11pt]{article}

\usepackage[]{acl}

\usepackage{times}
\usepackage{latexsym}
\usepackage{multirow}
\usepackage{comment}
\usepackage{multicol}
\usepackage{multirow}
\usepackage{booktabs}
\usepackage{xspace}
\usepackage{cleveref}
\usepackage{tabularx}
\usepackage[linewidth=1pt]{mdframed}
\usepackage{graphicx}
\usepackage{subcaption}

\newcolumntype{P}[1]{>{\centering\arraybackslash}p{#1}}

\usepackage[T1]{fontenc}

\usepackage[utf8]{inputenc}

\usepackage{microtype}

\definecolor{c1}{HTML}{4e79a7}%
\definecolor{c2}{HTML}{f28e2b}%
\definecolor{c3}{HTML}{009E73}%
\definecolor{c4}{HTML}{56B4E9}%
\definecolor{c5}{HTML}{CC79A7}%
\definecolor{c6}{HTML}{E69F00}%
\definecolor{c7}{HTML}{844E4D}%
\definecolor{c8}{HTML}{2D512A}%

\definecolor{oorange}{HTML}{d95f02}
\definecolor{bblue}{HTML}{7570b3}
\definecolor{ggreen}{HTML}{1b9e77}
\definecolor{ppurple}{HTML}{e37fbb}
\definecolor{lgreen}{HTML}{9CD24A}
\definecolor{yyellow}{HTML}{FFD52D}
\definecolor{ggold}{HTML}{E1BC89}
\definecolor{ggray}{HTML}{AAAAAA}
\newcommand{\hl}[2]{\colorbox{#1!20}{{#2}}}

\usepackage{tikz}

\newcommand{\corpus}{SourceSum}

\newcommand{\ppltag}[1]{\hl{c1}{\textsubscript{\tt Perplexity\textsubscript{#1}}}}
\newcommand{\rougetag}[1]{\hl{oorange}{\textsubscript{\tt ROUGE\textsubscript{#1}}}}
\newcommand{\attentiontag}[1]{\hl{yyellow}{\textsubscript{\tt Cross-attention\textsubscript{#1}}}}
\newcommand{\bertscoretag}[1]{\hl{ggold}{\textsubscript{\tt BERTScore\textsubscript{#1}}}}

\title{Source Identification in Abstractive Summarization}

\author{Yoshi Suhara\thanks{~~Work done while at Grammarly.}\\
  NVIDIA \\
  \texttt{ysuhara@nvidia.com} \\\And
  Dimitris Alikaniotis \\
  Grammarly \\
  \texttt{dimitris.alikaniotis@grammarly.com} \\}

\begin{document}
\maketitle

\begin{abstract}
Neural abstractive summarization models make summaries in an end-to-end manner, and little is known about how the source information is actually converted into summaries. 
In this paper, we define input sentences that contain essential information in the generated summary as {\em source sentences} and study how abstractive summaries are made by analyzing the source sentences. 
To this end, we annotate source sentences for reference summaries and system summaries generated by PEGASUS on document-summary pairs sampled from the CNN/DailyMail and XSum datasets. 
We also formulate automatic source sentence detection and compare multiple methods to establish a strong baseline for the task. 
Experimental results show that the perplexity-based method performs well in highly abstractive settings, while similarity-based methods perform robustly in relatively extractive settings.\footnote{Our code and data are available at \url{https://github.com/suhara/sourcesum}.} 
\end{abstract}

\section{Introduction}
Text summarization research has enjoyed recent advances in neural networks and pre-trained language models, which make abstractive summarization the most common approach~\cite{liu-lapata-2019-text,rothe-etal-2020-leveraging,pmlr-v119-zhang20ae}.
While continuing efforts in improving factuality and faithfulness~\citep{kryscinski-etal-2020-evaluating,nan-etal-2021-improving} have been made, abstractive summarization models, when trained properly, can create concise and coherent summaries from source documents.

Different from extractive summaries, for which we know the source information, it is not clear how an abstractive summary gathers various pieces of information that spread over different sentences in the input document (or input documents for multi-document summarization). 
Identifying source information is essential for the explainability and interpretability of summaries.

Therefore, in this paper, we aim to disentangle the abstractive summarization mechanism by identifying sentences that contain essential source information described in the generated summary. 
Existing studies use lexical similarity (e.g., ROUGE) and semantic similarity (e.g., BERTScore) for detecting sentences in the input document~\citep{vig-etal-2021-summvis, syed-etal-2021-summary} to help understand what the key source information for a generated summary.
Another line of work analyzes cross-attention weights for abstractive summarization~\citep{Baan:2019:UnderstandingMultiheadAttention} and data-to-text generation~\citep{juraska-walker-2021-attention}.
However, the approach mostly focuses on lexical and semantic similarity between the generated summary and input sentences without considering which input sentences provide source information.

To this end, we define input sentences that contain essential information for the generated summary as {\em source sentences} and aim to understand how abstractive summaries are composed by analyzing source sentences. 
We annotate source sentences for both reference summaries and system summaries generated by PEGASUS~\cite{Zhang:2020:PEGASUS} on the XSum and CNN/Daily Mail (CNN/DM) datasets, which are among the most popular summarization benchmarks in English. 
We also formulate the automatic source sentence detection task to verify the effectiveness of existing methods (i.e., attention-based and similarity-based) for detecting source sentences.
We develop a simple-yet-effective method based on perplexity gain---the difference in perplexity between the original text and the text after a specific sentence has been removed. We show that it significantly outperforms the existing methods in abstractive settings.

The contributions of the paper are as follows:
\begin{itemize}
    \setlength{\parskip}{0cm}
    \setlength{\itemsep}{0cm}
    \item We propose the novel task of automatic source sentence detection and create \corpus, which annotates source sentences of reference summaries and system summaries generated by PEGASUS on document-summary pairs sampled from XSum and CNNDM. 
   \item We develop a simple-yet-effective perplexity gain method to detect source sentences and 
   report that in a more abstractive setting, the perplexity gain method performs well while similarity-based methods can be a solid solution to extractive settings. 
\end{itemize}

\section{\corpus}
In this paper, we used XSum\footnote{\url{https://huggingface.co/datasets/xsum}}~\citep{narayan-etal-2018-dont} and CNN/DM\footnote{\url{https://huggingface.co/datasets/cnn_dailymail} version: 3.0.0
}~\citep{see-etal-2017-get} as the source datasets, as (1) they are the most common summarization benchmarks and (2) they have different levels of abstractiveness~\citep{narayan-etal-2018-dont}, to make the benchmark comprehensive and robust.

\subsection{Corpus creation}\label{subsec:corpus}
For each dataset, we randomly sampled document-summary pairs. 
We used a commonly used summarization model PEGASUS~\cite{Zhang:2020:PEGASUS} fine-tuned on either of the datasets.

In addition to generated summaries, we collect annotations for document-reference-summary pairs for the same set of examples, as abstractive summarization models may cause hallucinations, which would affect the quality of the benchmark. This setting also enables us to conduct a comparative analysis of reference and generated summaries.

\begin{figure}[t]
\includegraphics[width=0.49\textwidth]{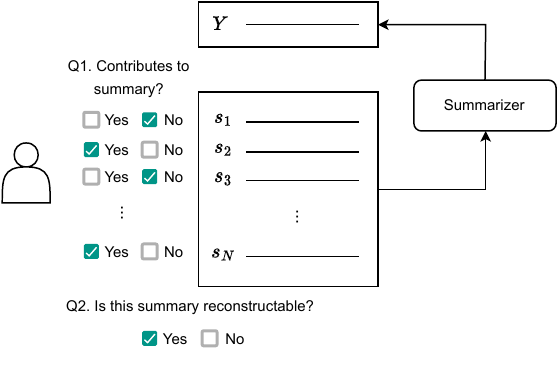}
    \caption{Annotation flow for \corpus. For each document-summary pair, the human annotator is asked to annotate each sentence (Q1), followed by the reconstructability question (Q2).%
    }\label{fig:annotation}
\end{figure}

\paragraph{Souce sentence annotation}
For each document-summary pair, the annotator is asked to judge if each sentence contributes to the summary after reading the summary and document (Q1 in Figure~\ref{fig:annotation}). The judgment criteria are whether the sentence (1) {\bf contributes to summary:} This sentence would be valuable in writing the summary, or (2) {\bf does not contribute to summary:} The summary could be written without this sentence.

\paragraph{Reconstructability annotation}
After completing the source sentence annotation step, the annotator was asked to answer a question ``Could you write this summary based solely on the sentences that you identified as important?'' to flag hallucinated summaries and ensure that \corpus{} consists of self-contained document-summary pairs.

This step is important for document-reference-summary pairs as well. As the reference summaries were taken from the introductory sentence (XSum) and the summary bullets (CNN/DM) of each article, it is not ensured that the reference summaries can be created solely from the original article, as reported in \citet{wang-etal-2020-asking}.

\subsection{Dataset Statistics}
We hired expert annotators to annotate source sentences on 2,000 document-summary pairs from XSum and CNN/DM. 
The inter-annotator agreement ratios (Krippendorff’s alpha) for the reconstructability annotation and source sentence annotation are 0.8 and 0.8, respectively.
As shown in Table~\ref{tab:reconstructable} and somewhat surprisingly, more than half of XSum summaries are not reconstructable, while most CNN/DM summaries are. After removing document-summary pairs that were judged non-reconstructable, \corpus{} consists of 1,211 document-summary pairs.

The basic statistics of \corpus{} are shown in Table~\ref{tab:stats}. Note that the summary is split into sentences for statistics calculation for the CNN/DM\footnote{XSum only contains single-sentence summaries.}.
The novel n-gram statistics show that PEGASUS generates quite extractive summaries (e.g., 2.9\% of unique unigram in generated summaries) for CNN/DM while generated summaries are still more abstractive for XSum.
This indicates that the behaviors of the two PEGASUS models fine-tuned on XSum and CNN/DM are different with respect to the abstractiveness of the generated summary.

\begin{table*}
\footnotesize
\begin{tabular}{l|ccccc|cccc}
\toprule
 &           &          &             &           &          & \multicolumn{4}{c}{\% of novel n-grams in summary} \\
 \corpus &  \# pairs &  \# sent & \# src sent & Input len & Summ len & unigram & bigram & trigram & 4-gram \\
\midrule
 XSum\textsubscript{PEGASUS}   & 119 & 10.28 & 3.09 (30.1\%) & 275.09 & 19.51 & 24.26 & 73.54 & 88.90 & 94.30 \\
 XSum\textsubscript{Reference} & 119 & 10.28 & 3.40 (33.1\%) & 275.09 & 23.71 & 33.93 & 82.54 & 94.10 & 97.58 \\
 CNN/DM\textsubscript{PEGASUS}   &  468 & 11.58 & 1.72 (14.9\%) & 309.07 & 16.95 &  2.90 & 19.26 & 29.96 & 37.20 \\
 CNN/DM\textsubscript{Reference} &  505 & 11.56 & 2.03 (17.6\%) & 305.79 & 15.87 & 13.53 & 50.45 & 67.92 & 77.02 \\ 
\bottomrule
\end{tabular}
\caption{Statistics of \corpus. Input len and Summ len are token counts using the PEGASUS tokenizer.
}\label{tab:stats}
\end{table*}

\section{Source Sentence Detection}

\paragraph{Problem Formulation}
Given an input document $X$, which consists of $N$ sentences $(s_1, \dots, s_N)$, and a system summary $Y$ generated by a summarization model $\theta$, the task is to identify a proper
subset of input sentences $D'$ that are essential to creating $Y$. The task can be cast as a sentence-scoring problem, where the score of each input sentence $R(s)$, assuming the threshold value $d$ to be a hyperparameter (i.e., $D' = \{s \in D|R(s) > d\}$).

\subsection{Similarity-based Method}
A simple approach is to choose sentences based on the similarity between the summary and input sentences.
The idea has been implemented in \citet{vig-etal-2021-summvis,syed-etal-2021-summary}, which use ROUGE and BERTScore for the similarity calculation. ROUGE puts more emphasis on lexical similarity while BERTScore takes semantic similarity into account. 

\begin{equation}
R(s, Y) = {\rm sim}(s, Y)
\end{equation}

Note that the similarity-based method is input- and model-agnostic, and it does not use $X$ and $\theta$ for relevance score calculation.
We also tested more sophisticated methods SimCSE~\cite{gao-etal-2021-simcse} and a PMI-based extractive summarization method~\citep{padmakumar-he-2021-unsupervised}, in addition to GPT-3.5 (\texttt{text-davinci-003}) \citep{InstructGPT:NeurIPS2022}. The prompt used for GPT-3.5 can be found in Appendix (Table~\ref{tab:gpt35_prompt}). 

We also used LexRank~\citep{erkan2004lexrank} as another baseline, as it can be used as a sentence-scoring method based on the centrality of the input sentence graph (i.e., summary-agnostic).

\subsection{Cross-attention Weights}
As the decoder takes input information via cross-attention, one approach is to calculate the importance of each sentence using cross-attention weights~\cite{juraska-walker-2021-attention}:
\begin{equation}
R(s, Y|X; \theta) =  \frac{1}{|s| |Y|} \sum_{x \in s} \sum_{y \in Y} w(x, y; \theta),
\end{equation}
where $w(x, y; \theta)$ denotes the cross-attention weight of the attention vector for the token $x$ in the encoder against the token $y$ in the decoder.
As the decoder typically has multiple attention heads on multiple Transformer layers, we calculate the average over the multiple heads and layers.

\subsection{Perplexity Gain}
Different from the similarity-based method, the attention-based method is {\em model-specific}, but is still an indirect method. 
Therefore, we consider a more direct way to calculate the importance of each sentence based on {\em perplexity gain} after removing the sentence:
\begin{equation}
R(s, Y|X; \theta) = {\rm PPL}(Y|X_{\setminus s}; \theta) - {\rm PPL}(Y|X; \theta),
\end{equation}
where $\mathrm{PPL}(Y|X; \theta)$ denotes the perplexity of the summary $Y$ generated by the model $\theta$ given the input document $X$. 
The intuition behind this method is that the model should be {\em more perplexed} (i.e., less confident) to generate the same summary if more relevant sentence is removed.

\section{Evaluation}

\paragraph{Evaluation metrics}
To make the evaluation independent of the choice of threshold selection, we used ranking metrics for evaluation, namely %
NDCG and MAP~\citep{IIR2008}. For NDCG, we used the total votes as the score to consider sentences with more votes more important. For MAP calculation, we binarized annotations and considered source sentences if two annotators agree it is relevant.

\paragraph{Results}
As shown in Table~\ref{tab:main}, Perplexity Gain outperforms the other methods for the XSum dataset, whereas the similarity-based methods perform best on the CNN/DM-Pegasus (SimCSE, BERTScore) and CNN/DM-Reference (ROUGE). 
The results confirm our hypothesis on the abstractiveness of summaries that it is necessary to access the summarization model for source identification.

\begin{table*}[t]
\footnotesize
    \centering
    \begin{tabular}{c|cc|cc|cc|cc}
    \toprule
     & \multicolumn{2}{c}{XSum\textsubscript{PEGASUS}} & 
     \multicolumn{2}{c}{XSum\textsubscript{Ref}} & 
     \multicolumn{2}{c}{CNN/DM\textsubscript{PEGASUS}} & 
     \multicolumn{2}{c}{CNN/DM\textsubscript{Ref}} \\
     & NDCG & MAP & NDCG & MAP & NDCG & MAP & NDCG & MAP \\\midrule
     LexRank~\citep{erkan2004lexrank} &  .7499 & .5302 & .7687 & .5435 & .6596 & .4226 & .6841 & .4540 \\ \midrule
     BERTScore~\citep{syed-etal-2021-summary} & .8499 & .6878 & .8762 & .7312 & .9134 & {\bf .8536} & .8851 & .7926 \\
     ROUGE~\citep{vig-etal-2021-summvis} & .8475 & .6740 & .8523 & .6756 & .9110 & .8484 & {\bf .8984} & {\bf .8087} \\ 
     SimCSE~\citep{gao-etal-2021-simcse} & .8579 & .7016 & .8661 & .7093 & {\bf .9141} & .8469 & .9048 & .8169 \\
     PMI~\citep{padmakumar-he-2021-unsupervised} & .8193 & .6316 & .8329 & .6480 & .8069 & .6919 & .7353 & .5592 \\ 
     GPT-3.5~\citep{InstructGPT:NeurIPS2022} &  .8233 & .5405 & .8422 & .5764 & .8095 & .5039 & .8252 & .5561 \\ \midrule
     Cross-attention~\citep{juraska-walker-2021-attention} &  .7048 & .4757 & --- & --- & .6312 & .3544 & --- & --- \\ \midrule
     Perplexity Gain & {\bf .8976} & {\bf .7753} & {\bf .8983} & {\bf .7710} & .8798 & .8138 & .8570 & .7465 \\
     \bottomrule
    \end{tabular}
    \caption{Performance of the source sentence detection methods on \corpus.}
    \label{tab:main}
\end{table*}

\begin{table*}[t]
\footnotesize
    \centering
    \begin{tabular}{cc|ccc|ccc}\toprule
           &       & \multicolumn{3}{c}{XSum} & \multicolumn{3}{c}{CNN/DM} \\
    Model & Input &      R1 &      R2 &      RL & R1 & R2 & RL \\\midrule
    \multirow{2}{*}{PEGASUS} & All sentences &  53.40 &  30.49 &  45.38 & 47.13 &  25.75 &  35.54 \\
     & Source sentences only & {\color{red}48.36$\downarrow$} &  {\color{red}25.62$\downarrow$} &  {\color{red}40.44$\downarrow$} & {\color{ggreen}47.55$\uparrow$} &  {\color{red}25.68$\downarrow$} & {\color{ggreen}36.16$\uparrow$} \\ \midrule
    \multirow{2}{*}{BART} & All sentences &  50.32 &  26.35 & 40.83 & 45.56 & 23.32 & 32.70 \\
     & Source sentences only & {\color{red}47.29$\downarrow$} &  {\color{red}22.73$\downarrow$} &  {\color{red}38.82$\downarrow$} & {\color{ggreen} 47.53 $\uparrow$} &  {\color{ggreen} 24.92 $\uparrow$} & {\color{ggreen} 34.58 $\uparrow$} \\\midrule    
    \multirow{2}{*}{LexRank} & All sentences &  19.84 &  3.08 &  14.46 & 37.30 &  15.94 &  23.45 \\
    & Source sentences only &  {\color{ggreen}23.36$\uparrow$}  &  {\color{ggreen}5.74$\uparrow$} &  {\color{ggreen}17.49$\uparrow$} & {\color{ggreen}45.45$\uparrow$} &  {\color{ggreen}23.47$\uparrow$} & {\color{ggreen}27.04$\uparrow$}  \\ \bottomrule
    \end{tabular}
    \caption{Summarization performance of PEGASUS, BART, and LexRank on \corpus{} (XSum and CNN/DM). Using only source sentences as input improves LexRank's performance on both datasets, while significant degradation is observed for PEGASUS and BART on XSum.}
    \label{tab:sourcesent}
\end{table*}

\section{Analysis}
\noindent
{\bf Are summaries reconstructable?} 
As reference summaries for the XSum (CNN/DM) dataset were scraped from the introductory sentence (the summary bullets), it is not ensured that reference summaries can be created only from the input documents.
The same thing can be said for summaries generated by abstractive summarization models, which may hallucinate content. 
To analyze this, we annotated document-summary pairs with respect to the reconstructability (\S\ref{subsec:corpus}). 

Table~\ref{tab:reconstructable} shows that more than half of XSum summaries are not reconstructable, while most of CNN/DM summaries are. 
Compared to the reference summaries, summaries generated by the Pegasus models are slight more reconstructable, as expected.
The higher reconstructability of CNN/DM is also supported by the lower abstractiveness (i.e., lower novel $n$-grams).

\begin{table}[t]
\footnotesize
    \centering
    \begin{tabular}{c|cc|cc}    
    \toprule  
    \multirow{2}{3.5em}{Reconst\-ructable?}
    & \multicolumn{2}{c|}{\textbf{XSum}} & \multicolumn{2}{c}{\textbf{CNN/DM}} \\
     & Ref. & PEGASUS & Ref. & PEGASUS \\\midrule
    Yes & 30.3\% & 37.3\% & 87.7\% & 95.0\% \\
    Partly & 18.1\% & 15.4\% & 4.1\% & 3.0\% \\
    No & 51.7\% & 47.3\% & 8.2\% & 2.0\% \\\bottomrule
    \end{tabular}
    \caption{reconstructability of reference/generated summaries. More than half of XSum reference summaries cannot be created only from the input document.}
    \label{tab:reconstructable}
\end{table}

\begin{figure}[t]
    \centering
     \begin{subfigure}[b]{0.23\textwidth}
         \centering
         \includegraphics[width=\textwidth]{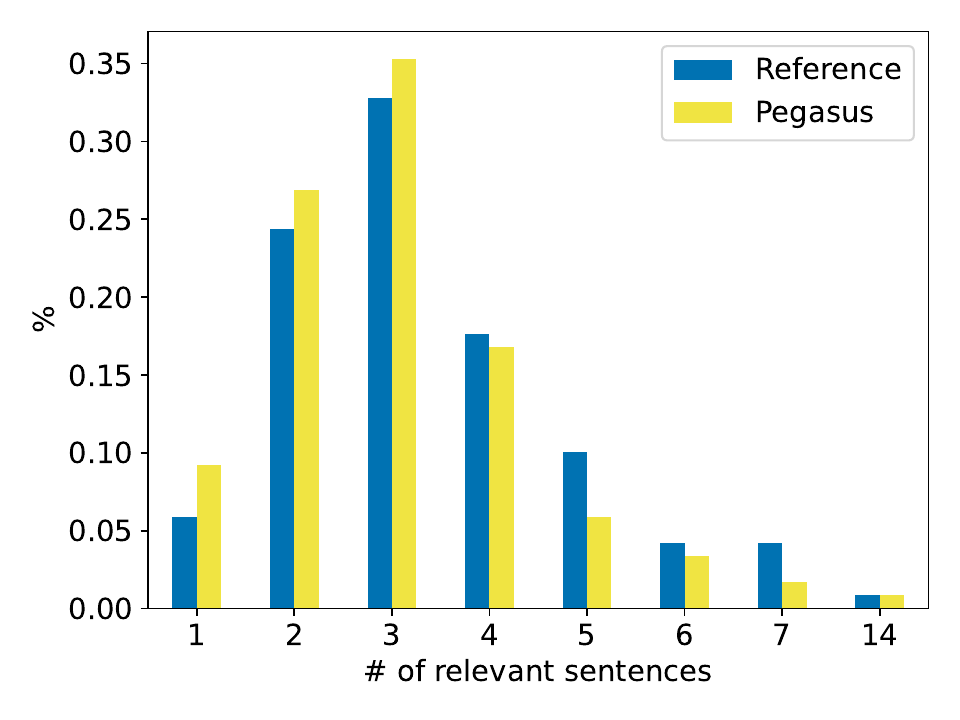}
         \caption{XSum}
     \end{subfigure}
     \hfill
     \begin{subfigure}[b]{0.23\textwidth}
         \centering
         \includegraphics[width=\textwidth]{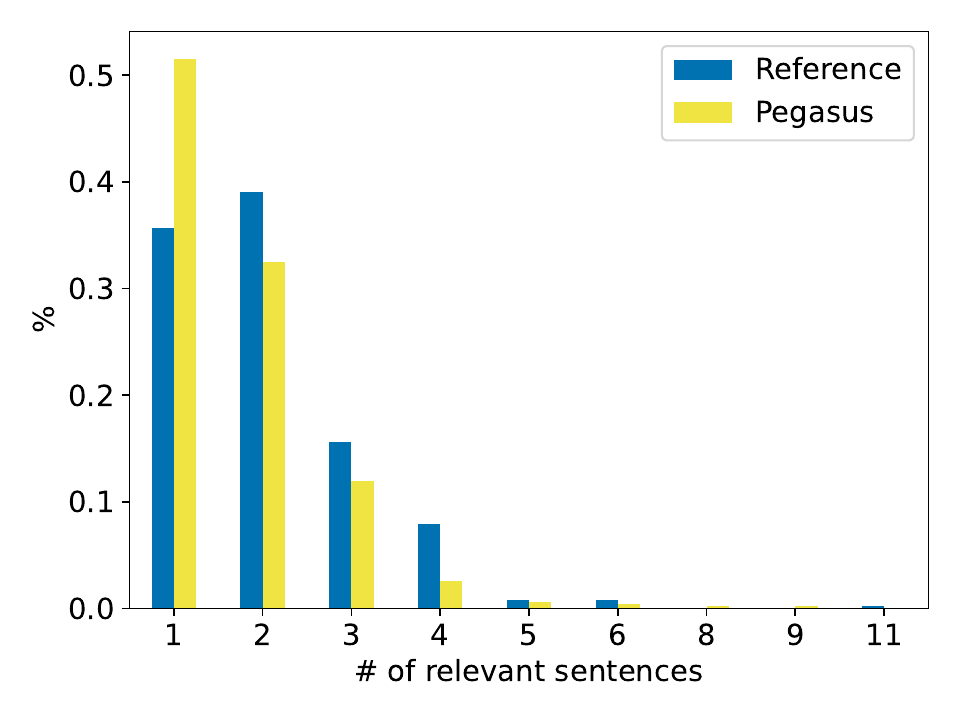}
         \caption{CNN/DM}
     \end{subfigure}
     \caption{Distribution of the number of (ground-truth) source sentences. Generated summaries tend to have more source sentences on XSum while having fewer source sentences on CNN/DM.%
     }\label{fig:num_src_sent}
\end{figure}

\noindent
{\bf How many source sentences are used per summary?} 
Figure~\ref{fig:num_src_sent} shows the distribution of the number of source sentences per one summary sentence. As shown in the figure, XSum summaries have more source sentences (3.40 on average) than CNN/DM summaries (1.72 on average). The trend is aligned with the abstractiveness/extractiveness of those datasets. 
Regarding the differences in reference and generated summaries, PEGASUS amplifies the characteristics of each dataset---Generated summaries tend to have more (less) source sentences on XSum (CNN/DM).

\paragraph{Are non-source sentences unnecessary?}
We have defined source sentences from which summaries can be made. A natural question is whether the other ``non-source'' sentences are necessary for generating the same abstractive summaries. To answer the question, we evaluated the quality of summaries by PEGASUS, BART~\cite{lewis-etal-2020-bart}, and LexRank under two settings: (1) All sentences and (2) source sentence only. 

Results are shown in Table~\ref{tab:sourcesent}.
Interestingly and somewhat surprisingly, by removing non-source sentences, PEGASUS and BART show significant degradations on XSum while slight improvements are observed on CNN/DM.
In fact, we confirm some degree of hallucinations when generating with source sentence only, as shown in Table~\ref{tab:sourcesent_summary}.
We consider that especially in an abstractive setting, non-source sentences still provide context information, which helps give {\em confidence} to the summarization model.
From the results, we confirm that abstractive summarization by the pre-trained Transformer model is more complicated than simply selecting and rewriting source information. 
The quality improvements for LexRank are reasonable as LexRank should be a higher chance to select relevant sentences in the source sentence-only setting.

\begin{figure}[t]
    \centering
     \begin{subfigure}[b]{0.23\textwidth}
         \centering
         \includegraphics[width=\textwidth]{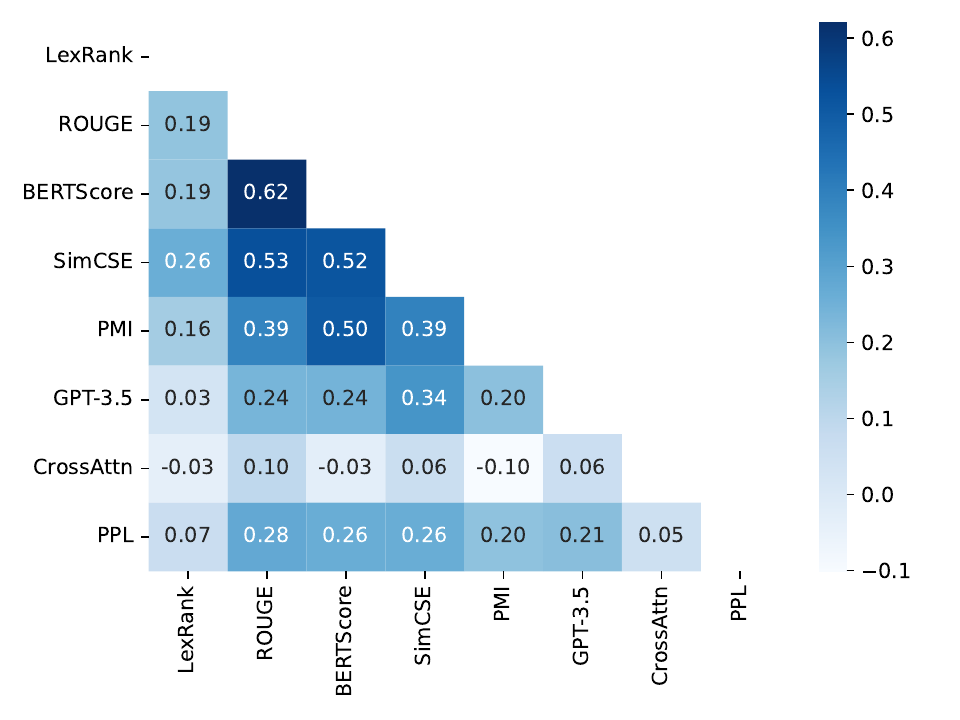}
         \caption{XSum}
     \end{subfigure}
     \hfill
     \begin{subfigure}[b]{0.23\textwidth}
         \centering
         \includegraphics[width=\textwidth]{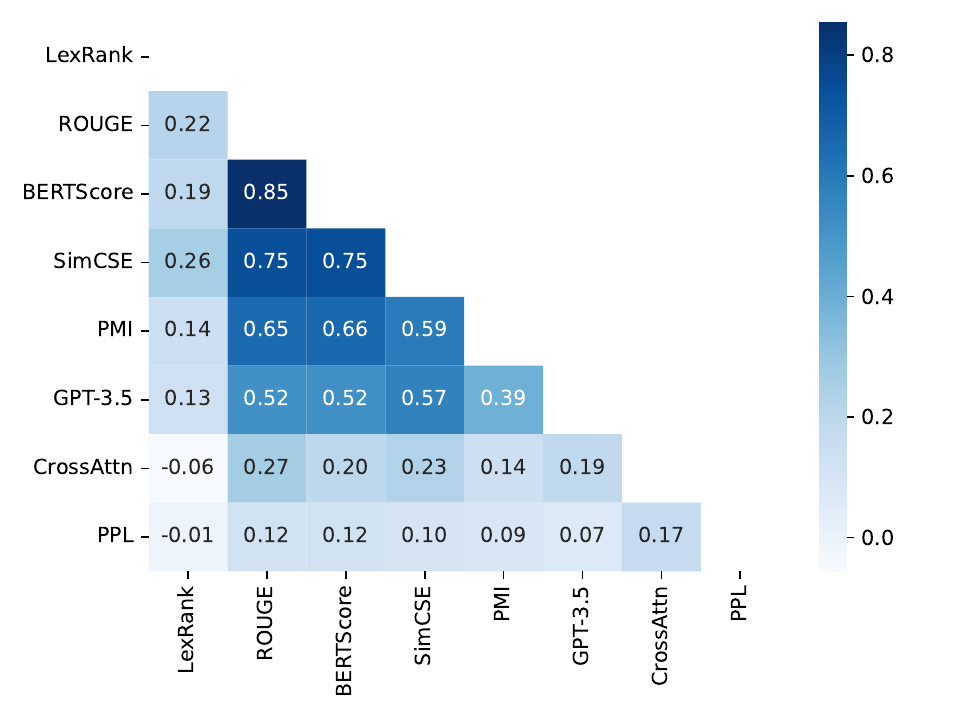}
         \caption{CNN/DM}
     \end{subfigure}    
    \caption{Correlation analysis of the source sentence detection methods. %
    }\label{fig:correlation}
\end{figure}

\paragraph{Do different methods detect different source sentences?}
Table~\ref{tab:main} does not show if different methods detect the same or different source sentences. To analyze this, we calculated correlation coefficients of scores calculated by the different methods.
Figure~\ref{fig:correlation} shows that the scores of the similarity-based methods 
are highly correlated 
while Perplexity Gain and Cross Attention detect source sentences differently.

\section{Related Work}

It is hard to interpret how commonly used Transformer-based summarization models generate abstractive summaries.
\citet{xu-durrett-2021-dissecting} developed an ablation-attribution framework that identifies the generation model by comparing behaviors of a language model and a summarization model. 
\citet{Baan:2019:UnderstandingMultiheadAttention} investigated the interpretability of multi-head attention in abstractive summarization and found that attention heads can be pruned without a significant performance drop.

Another line of work analyzes how multiple sentences are fused into summary sentences~\citep{lebanoff-etal-2019-analyzing,lebanoff-etal-2019-scoring, lebanoff-etal-2020-learning,lebanoff-etal-2020-understanding}. \citet{lebanoff-etal-2020-understanding} created a dataset that contains fine-grained point-of-correspondence between a summary and two source sentences. Our work covers beyond the scope of their work as \corpus{} assigns source sentence labels to all source sentences on both generated and human summaries.

One simple-yet-effective approach for explainability is to highlight sentences similar to the generated summary. \citet{vig-etal-2021-summvis} and \citet{syed-etal-2021-summary} use ROUGE and BERTScore to capture the lexical and semantic similarity to help the user understand the source information of the generated summary. \citet{juraska-walker-2021-attention} use cross-attention to understand the behavior of the data-to-text model. \citet{Wang2021ExploringES} develops a hybrid summarization model that takes into account sentence similarity to improve explainability and faithfulness. 
\citet{saha2023summarization} develops a framework that uses neural modules to construct a tree representation to understand the relationship between a human-written summary and the input document. 
This paper is aligned with the line of work but rather focuses on formulating the source sentence detection task and creating a benchmark, so we can evaluate and compare different methods quantitatively and qualitatively.

\section{Conclusion}
In this paper, we formulate the source sentence detection task, which finds input sentences that are essential to generating the given abstract summary, to study how abstractive summaries are made.
We annotated source sentences for reference summaries and system summaries generated by PEGASUS on XSum and CNN/DM and created a benchmark \corpus.
Experimental results on \corpus{} show that Perplexity Gain, which is based on the perplexity increase when the target sentence is removed, performs the best in highly abstractive settings (XSum), while similarity-based methods perform robustly in extractive settings (CNN/DM).

\section*{Limitations}
As we shed light on a new perspective on abstractive summarization, the paper has certain limitations. First, our benchmark \corpus{} is made for single-document summarization in a single domain (news) in a single language (English), which not necessarily ensuring the generalizability for other domains and languages. For multi-document summarization, we believe that the same annotation and evaluation framework can be applied and is interesting future work.
Second, the annotation is sentence-level in \corpus. There may be a chance that annotated source sentences also contain information unnecessary to generate the summary. We carefully discussed the annotation guideline and decided to use sentence-level annotation to ensure the annotation quality.
Last but not least, the benchmark is created on top of a Transformer-based encoder-decoder model PEGASUS and the results do not necessarily apply to other encoder-decoder models or autoregressive models such as GPT series. 
With those limitations, we still believe that the paper and the benchmark are beneficial for the community in providing insights into abstractive summarization models.

\section*{Acknowledgements}
We thank Jennifer Bellik for her support on the annotation collection and thank the anonymous reviewers for their helpful comments.

\bibliography{anthology,custom}
\bibliographystyle{acl_natbib}

\appendix

\section{Annotation collection}\label{app:annotation}

\subsection{Data preparation}
Following the official script used to fine-tune the summarization models, we filtered out examples whose number of tokens in the input document is greater or less than certain numbers.

\noindent
{\bf XSum} We sampled document-summary pairs from the XSum dataset\footnote{\url{https://huggingface.co/datasets/xsum}}. We filtered examples whose number of tokens is greater than 56 and less than 512. %

\noindent
{\bf CNN/DM}  We sampled document-summary pairs from the CNN/DailyMail dataset\footnote{\url{https://huggingface.co/datasets/cnn_dailymail} 3.0.0}. We filtered examples whose number of tokens is greater than 142 less than 1024. %

\subsection{Summary generation}
\noindent
{\bf XSum} {\tt pegasus-xsum}\footnote{\url{https://huggingface.co/google/pegasus-xsum}} with the default generation configuration (length\_penalty = 0.6, max\_length = 64, num\_beams = 8). 

\noindent
{\bf CNN/DM} {\tt pegasus-cnn\_dailymail}\footnote{\url{https://huggingface.co/google/pegasus-cnn_dailymail}} with the default generation configuration (length\_penalty = 0.8, min\_length = 32, max\_length = 128,  num\_beams = 8). Summaries are split by {\tt <n>} into sentences.

\subsection{Pilot Study} 
We conducted two pilot studies to revise the annotation guideline while helping the annotators familiar with the annotation task.  We initially used ternary labels (Essential, Related, Unrelated) for annotation. However, the inter-annotator agreement was not sufficiently high (Krippendorff’s alpha was 0.443).
Thus, we decided to use binary labels and further clarify the label definitions. %
Also, we decided to exclude input documents that consist of more than 15 sentences, based on the feedback from the annotators, to reduce the cognitive load and to ensure the annotation quality.

\subsection{Annotation guideline}
Figure~\ref{fig:annotation} depicts the annotation workflow. For each document-summary pair, the human annotator submits source sentence labels followed by a reconstructability label.
The full annotation guideline and reconstructability judgment guideline are
shown in Tables~\ref{tab:annotation_guideline} and \ref{tab:reconstructability_guideline}.

\section{Source Sentence Detection}
Table~\ref{tab:gpt35_prompt} is the prompt used for GPT-3.5 to obtain source-sentence scores.

\begin{table}[t]
    \centering
    \footnotesize
    \begin{tabular}{p{22.5em}}
        \toprule
This task is to identify the sentences in a document that contribute to a given summary of that document. This annotation is a sentence-labeling task. For each snippet, you'll see a summary (labeled Summary:) and a sentence of a short news article (labeled Sentence:).\\
\vspace{0.5em}
The output will be a score from 0 to 100,  0 with ``doesn't contribute to summary'' with the highest confidence and 100 with ``contribute to summary'' with the highest confidence.\\
\vspace{0.5em}
Summary: \texttt{\{summary\}}\\
Sentence: \texttt{\{sentence\}}\\
\vspace{0.5em}
Score: \\ \bottomrule
    \end{tabular}
    \caption{Prompt for GPT-3.5 used in the experiment.}
    \label{tab:gpt35_prompt}
\end{table}

\section{Analysis}
In this section, we report a more detailed analysis on \corpus.

\begin{figure}[t]
    \centering
     \begin{subfigure}[b]{0.23\textwidth}
         \centering
         \includegraphics[width=\textwidth]{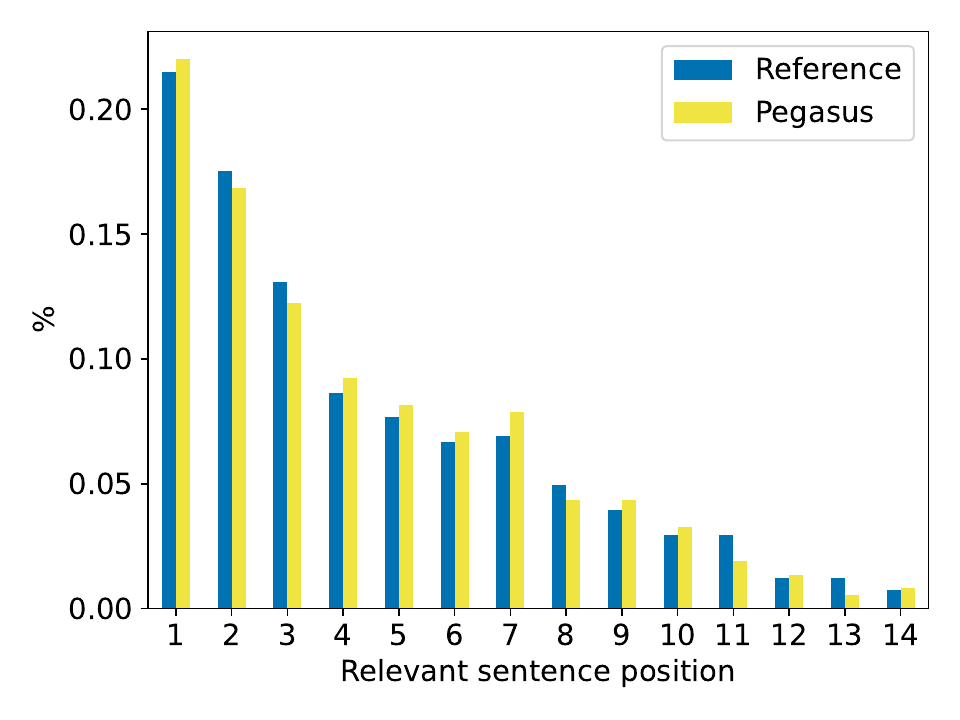} 
         \caption{XSum}
     \end{subfigure}
     \hfill
     \begin{subfigure}[b]{0.23\textwidth}
         \centering
         \includegraphics[width=\textwidth]{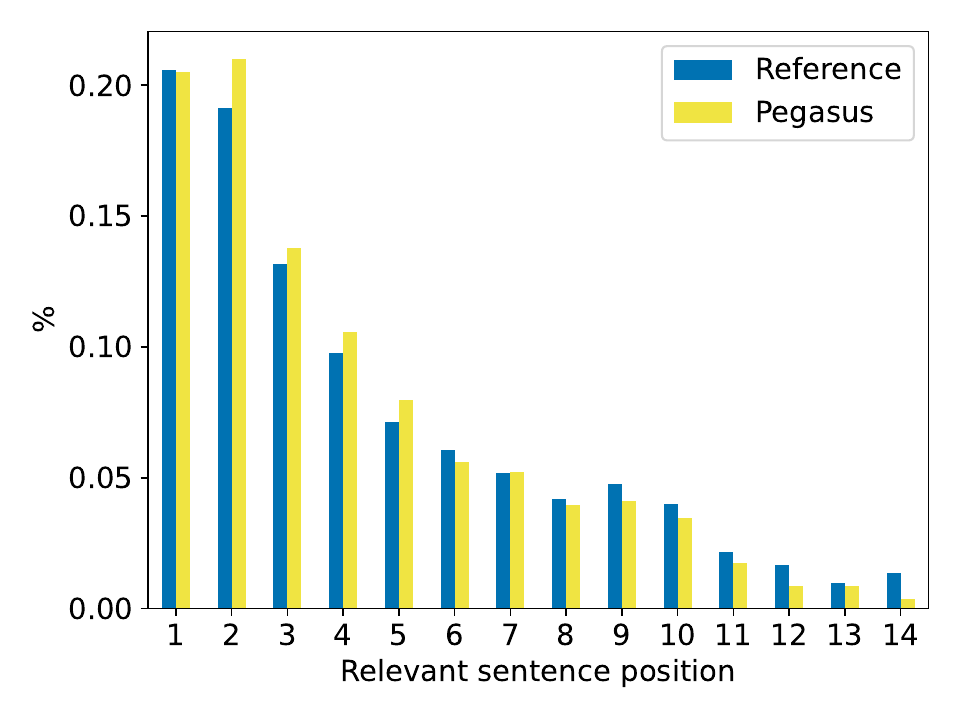}
         \caption{CNN/DM}
     \end{subfigure}
     \caption{Distribution of source sentence absolute positions. Both plots support that a commonly used lead-3 .}\label{fig:abs_pos}
\end{figure}

\begin{figure}[t]
    \centering
     \begin{subfigure}[b]{0.23\textwidth}
         \centering
         \includegraphics[width=\textwidth]{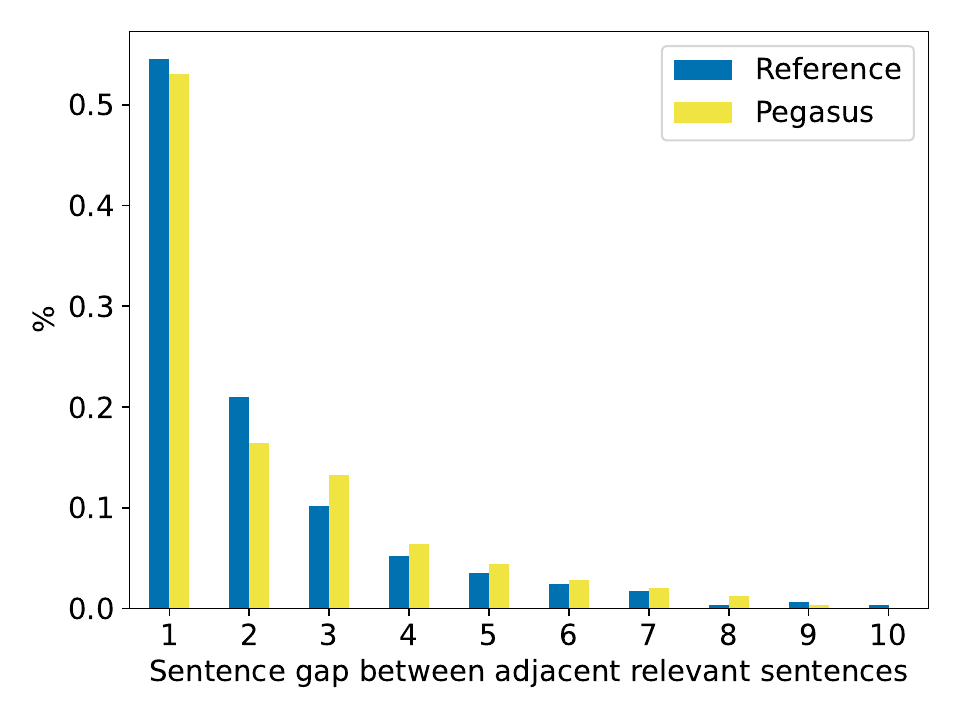}
         \caption{XSum}
     \end{subfigure}
     \hfill
     \begin{subfigure}[b]{0.23\textwidth}
         \centering
         \includegraphics[width=\textwidth]{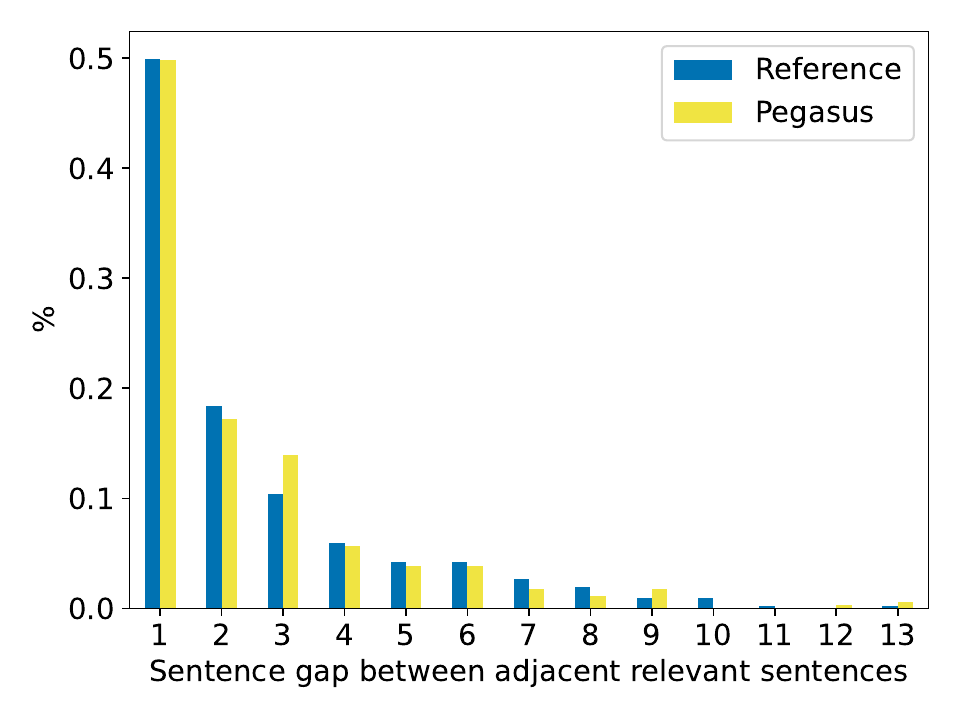}
         \caption{CNN/DM}
     \end{subfigure}
     \caption{Distribution of the sentence interval between adjacent (ground-truth) source sentences. For example, if source sentence positions are 1, 3, and 7, the sentence intervals for the example are 2 and 4.  }\label{fig:rel_pos}
\end{figure}

\subsection{Source sentence distribution}
Figure~\ref{fig:abs_pos} shows the sentence positions of source sentences. As expected, source sentences tend to appear at the beginning of the document, which supports the idea of using lead sentences as simple-yet-effective heuristics for news summarization~\citep{Zhu:2021:SIGIR:LeadBias}. The position bias has also been reported in \citep{kryscinski-etal-2019-neural} and \citep{LeveragingLeadBias::SIGIR2021}.
However, the plots also show that source sentences spread over the document, which indicates that summarization involves more complex textual processing.

The sentence intervals between adjacent source sentences follow a similar distribution on XSum and CNN/DM. 
Figure~\ref{fig:rel_pos} shows that source sentences generally distribute closely in the source document.

\subsection{Qualitative Analysis}
Table~\ref{tab:qualitative} shows ground truth and detected source sentences for a summary.
Ground-truth source sentences are highlighted in \hl{ggreen}{green} and the top-$k$ results by each method are tagged at the end of each sentence. In this examples, we highlight the same number of source sentences as the number of ground-truth source sentences (i.e., $k=2$ in the table). 
In this example, only Perplexity Gain successfully detected (S1) and (S8) as the source sentences for the summary.

\begin{table*}[t]
    \centering
    \footnotesize
    \begin{tabular}{p{\dimexpr\linewidth-10pt}}
        \toprule
        \textbf{Summary:} The Nigg Energy Park in Ross-shire has been awarded a contract to assemble offshore wind turbines. \\\midrule
        \textbf{Input document:} 
        \hl{ggreen}{(S1) The site owned by Global Energy Group joins Wick Harbour in Caithness in securing work on the} \hl{ggreen}{ Â£2.6bn Beatrice Offshore Windfarm Ltd (Bowl) project.}\ppltag{1}
(S2) Siemens, one of the companies involved in Bowl, will use the yard for assembling turbines from spring 2018.
(S3) Once assembled the turbines would be towed out to the wind farm site.\rougetag{2}\bertscoretag{2}
(S4) The project, which also involves energy giant SSE, is to be created about eight miles off Wick.
(S5) Global said Nigg's involvement would help to secure work for more than 100 people.
(S6) The Scottish government, Highland Council, Highlands and Islands Enterprise, Scottish Council for Development and Industry (SCDI) and Scottish Renewables have welcomed the announcement.
(S7) Business, Innovation and Energy Minister, Paul Wheelhouse, said: "Offshore renewables represent a huge opportunity for Scotland; an opportunity to build up new industries and to deliver on Scotland's ambitious renewable energy and carbon reduction targets for 2020 and beyond.
\hl{ggreen}{(S8) "I am delighted that this multi-million pound contract between Global Energy Group and Siemens will enable Nigg} \hl{ggreen}{Energy Park to develop into a genuine multi-energy site, securing around 100 direct and indirect jobs and associated supply} \hl{ggreen}{chain opportunities.}\ppltag{2}
(S9) "This contract arising from installation of the Beatrice Offshore Wind farm will provide a very welcome boost to the local economy in Ross-shire and the wider Highland Council area."\rougetag{1}\bertscoretag{1}
(S10) Regional director for the Highlands and Islands, Fraser Grieve, said: "Today's announcement of Nigg's involvement in the Beatrice Offshore wind project shows the positive economic impact that this major development will have on the region over the coming years.\attentiontag{2}
(S11) "Nigg, and the wider Cromarty Firth, has much to offer and this agreement is not only a boost for the Global Energy Group but will benefit the supply chain through the area."
(S12) Lindsay Roberts, senior policy manager at renewable energy industry group Scottish Renewables, said: "The contract signed today will help breathe new life into this Highland port.
(S13) "Scotland's offshore wind industry has huge potential for both our economy and our environment, and it's great to see Nigg reaping the benefits.
(S14) "As other wind farms with planning consent in the Scottish North Sea begin to develop, agreements like this will play a key role in securing benefits not just for communities on the east coast, but for the whole of Scotland."\attentiontag{1} \\\bottomrule
    \end{tabular}
    \caption{Output examples of the source sentence detection methods. The source sentences are highlighted in \hl{ggreen}{green}. Tag(s) appended to the end of a sentence denote the method names and the ranks. In this example, only Perplexity Gain successfully detected (S1) and (S8) as the source sentences.}
    \label{tab:qualitative}
\end{table*}

\begin{table*}[t]
    \centering
    \footnotesize
    \begin{tabular}{p{\dimexpr\linewidth-10pt}}
        \toprule
        \textbf{Input document:}
        \hl{ggreen}{The Tories won 37 of 64 seats to claim a majority and wipe out Labour's 22-seat majority from 2013.}
Labour picked up 24 seats this time around, the Liberal Democrats won three and UKIP finished with none.
Towns where seats turned from red to blue included Swadlincote, Matlock, Glossop, Buxton, Ripley, Belper and Ilkeston.
Turnout was 38\%. Election 2017: Full results from across England Conservative leader Barry Lewis described the result as "brilliant". \hl{ggreen}{"We didn't} \hl{ggreen}{ think at this point in the electoral cycle we'd be taking control of Derbyshire County Council," he said.}
"We fought a really good campaign on local issues and I think that's really helped. We got our manifesto out early and really hit the doorsteps."
This was Labour's last stand - its last county council to be defended in England. And its defences have proven to be weak.
The Conservatives have won across the south and centre of the county - in places like Heanor, Ilkeston and Ripley.
They've also taken seats off the Lib Dems. And it was a bad night too for UKIP - their share of the vote in Derbyshire collapsed.
\\\midrule
        \textbf{Reference summary: } The Conservatives have taken control of Derbyshire County Council with a massive swing from Labour.\\\midrule
        \textbf{With all sentences (PEGASUS): } The Conservatives have taken control of Derbyshire County Council.\\\midrule
        \textbf{With source sentences only (PEGASUS): } Conservative leader \hl{ppurple}{Simon Danczuk} has said \hl{ppurple}{he is "delighted"} his party has taken control of Derbyshire County Council.\\
        \bottomrule
    \end{tabular}
    \begin{tabular}{p{\dimexpr\linewidth-10pt}}
        \toprule
        \textbf{Input document:}
        \hl{ggreen}{Stuart Campbell was arrested in the west of England on Friday following a complaint from a woman in} \hl{ggreen}{south London.} \hl{ggreen}{She had made allegations of harassment taking place over a two-year period.}
Mr Campbell, who was released on bail, said it concerned some tweets and insisted they were not threatening.  He accused the media of "innuendo" designed to encourage "speculations". \hl{ggreen}{The blogger, a former computer games reviewer who was born in Stirling but lives} \hl{ggreen}{in Bath, has been a vocal campaigner for Scottish independence and launched the Wings Over Scotland blog in 2011.}
On Friday he tweeted that he would be posting less frequently than usual because of "reasons totally outwith my control (don't ask)".
End of Twitter post  by @WingsScotland
\hl{ggreen}{A spokesman for the Metropolitan Police said: "Police are investigating an allegation of } \hl{ggreen}{online harassment.}
"The allegation was made after a woman, aged in her 30s, attended a south London police station. The harassment is said to have taken place over the past two years."
Mr Campbell has been bailed, pending further inquiries, to a date in mid-September.
In a statement on the Wings Over Scotland website, Mr Campbell responded to a report of his arrest which appeared in The Herald newspaper.
He said that piece "has been written for maximum innuendo to allow the wildest speculations on social media - which are of course duly taking place - but the alleged events relate entirely to some tweets from our Twitter account, none of which have been deleted and all of which are still publicly visible.
"Nothing more sinister or serious than some tweets has occurred or been alleged to have occurred. None of the tweets involved are in ANY way threatening, not even in a joking sense. That's all we'll be saying on the subject at this time."
\\\midrule
        \textbf{Reference summary: } The pro-independence blogger behind the Wings Over Scotland website has been arrested for alleged online harassment. \\\midrule
        \textbf{With all sentences (PEGASUS): } A pro-independence blogger has been arrested on suspicion of online harassment. \\\midrule
        \textbf{With source sentences only (PEGASUS): } A prominent Scottish independence blogger \hl{ppurple}{has been released without charge} after being arrested on suspicion of online harassment.\\
        \bottomrule
    \end{tabular}    
    \caption{Examples of summaries generated with all sentences and with source sentences only (XSum). The source sentences in the input document are highlighted in \hl{ggreen}{green}. Incorrect/hallucinated words are highlighted in \hl{ppurple}{purple}.}
    \label{tab:sourcesent_summary}
\end{table*}

\begin{table*}[t]
    \centering
    \footnotesize
    \begin{tabular}{p{\dimexpr\linewidth-10pt}}
        \toprule
        \textbf{Annotation guideline}\\\midrule
        
        \textbf{Goals:}
        Your task in this annotation is to provide the ``highlighting'' for document-summary pairs, and check the validity of summaries.

        \begin{enumerate}
          \setlength{\parskip}{0cm}
          \setlength{\itemsep}{0cm}           
        \item To identify the sentences in a document that contribute to a given summary of that document.
        \item To determine whether a given summary is valid (all the important points in it are captured in the document itself).
        \end{enumerate}

        \textbf{Instructions:}
        This annotation is a sentence-labeling task. For each snippet, you'll see a summary (labeled SUMMARY:) and a short news article (labeled DOCUMENT:).

        \vspace{1em}

        \textbf{Summary:}
        The summary appears in multiple places for each snippet in order to eliminate the need to scroll up and down. It is first shown before the document because it often functions as the first sentence of the article. Secondly, the summary appears in the {\bf Prompt} box to the right of the editable window, so that you can always refer to it without needing to scroll.
        \vspace{0.5em}

        Lastly, the summary appears at the bottom of the editable window, labeled SUMMARY: again. This final repetition is pre-tagged with the question Reconstructable? so that you can label it. As yourself, ``Could I reconstruct all the important points of this summary based only the sentences I labeled as `1: contributes'?'' and answer {\bf Yes, reconstructable} or {\bf No, not reconstructable}.

      \vspace{1em}
      \textbf{Document:}
      The document is pre-annotated with sentence-boundaries. The end of each sentence is tagged with the question 0 or 1?. Mark sentences that are important to the provided summary as {\bf 1: contributes to summary}. Mark sentences that are not important to the summary as {\bf 0: doesn't contribute to summary}.
      \vspace{0.25em}

      Documents in this annotation are either CNN (three fifths) or BBC (two fifths) news articles. Some summaries are written by the articles' authors, others are generated by models.

      \vspace{0.5em}
      \textbf{For Duplicates:}
      You will sometimes see the same document multiple times, paired with a different summary each time. This can happen for two reasons:

    \begin{enumerate}
          \setlength{\parskip}{0cm}
          \setlength{\itemsep}{0cm}       
    \item because we are considering multiple sources of summaries, and
    \item because original summaries for some articles were multiple sentences, and we are only displaying one summary sentence at a time.
    \end{enumerate}

    Each document-summary {\em pair} that you see should be unique, however.
    \vspace{1em}

    \textbf{Annotation steps:}

    \begin{enumerate}
      \setlength{\parskip}{0cm}
      \setlength{\itemsep}{0cm}        
      \item Read the summary at the top of the editing window, then read the document.
      \item Evaluate each sentence for whether it provides information that contributes to the summary. (You can refer to the summary in the prompt on the right if you've scrolled down from the summary in the editing window.) Label every sentence in the document with one of the following labels:
    \begin{itemize}
      \setlength{\parskip}{0cm}
      \setlength{\itemsep}{0cm}        
    \item {\bf 1: contributes to summary}: This sentence would be valuable in writing the summary.
    \item {\bf 0: doesn't contribute to summary}: The summary could be written without this sentence.
    \end{itemize}
    \end{enumerate}

    3. Now that you've read the document, assess whether the important points of the summary (repeated at the bottom of the document) are also present in the document itself. Answer the question, "Could you write this summary based solely on the sentences that you identified as important?"
    \begin{itemize}
      \setlength{\parskip}{0cm}
      \setlength{\itemsep}{0cm}        
    \item If so, label the summary at the bottom of the document with {\bf Yes, reconstructable}.
    \item If you would need additional information to write the summary, OR if the summary contradicts the document, then label it as {\bf No, not reconstructable}.
    \item You can also change the labels of sentences in the document if you realize that more of them are needed in order to write the summary.
    \end{itemize}

    4. When all sentences have been labeled and you've evaluated the summary, click "Submit" and review your annotations.
    \begin{itemize}
      \setlength{\parskip}{0cm}
      \setlength{\itemsep}{0cm}        
        \item Read over just the sentences that you marked as {\bf 1: contributes to summary}, and confirm that each of them contains information that the summary directly includes.
        \item If you labeled the summary as {\bf Yes, reconstructable}, verify that all the important   information in it is contained in the sentences marked {\bf 1: contributes}.
    \end{itemize}
    \\
    \bottomrule
    \end{tabular}
    \caption{Annotation guideline.}
    \label{tab:annotation_guideline}
\end{table*}

\begin{table*}[t]
    \centering
    \footnotesize
    \begin{tabular}{p{\dimexpr\linewidth-10pt}}
        \toprule
        \textbf{Reconstructability judgment guideline---Determining whether a summary is reconstructable}\\
        \midrule
        We'll count a summary as valid and reconstructable if all the {\bf important points} in it can be reconstructed from the document by a reader who is part of the document's {\bf intended audience}.
        \vspace{0.25em}
        
        What counts as an ``important point'' is somewhat subjective, but here is some guidance:

        \vspace{0.5em}

        \textbf{Important to be able to reconstruct from the document:}
        \begin{itemize}
          \setlength{\parskip}{0cm}
          \setlength{\itemsep}{0cm}        
            \item All named entities (e.g., Wales Under-20, Samoa, World Rugby U20 Championship, Georgia): If a name appears in the summary, it is an important point in the summary. Only mark the summary as reconstructable if the name or entity also appears in the document. It's okay if a co-referring expression (but not the exact name itself) appears in the document.
            \item Events
            \item Approximate quantities; exact values don't need to be reconstructable (e.g., ``10,000 free racquets'' in the summary could be supported by ``many free racquets'' in the document; ``1.9\% increase'' could be supported by ``about 2\% increase'')
        \end{itemize}

        \textbf{Not important to be able to reconstruct from the document:}
        \begin{itemize}
          \setlength{\parskip}{0cm}
          \setlength{\itemsep}{0cm}        
            \item {\em Expansions of acronyms or abbreviations}: If the full phrase that an acronym stands for appears in the summary but not in the document, the summary can still be considered reconstructable; the expansion of the acronym is a minor point in the summary, not an important point. Different expressions that refer to the person or place mentioned in the summary qualify as
            \item {\em Exact numbers} are not important. No need to break out the calculator.
            Information sources (e.g., “State television reports”, “Official figures show”).
        \end{itemize}

        \textbf{Some summary examples with important information in {\em italics}:}
        \begin{itemize}
          \setlength{\parskip}{0cm}
          \setlength{\itemsep}{0cm}
        \item {\em Wales Under-20} ran in {\em eight tries} to {\em beat Samoa} and secure their {\em first win} of {\em the World Rugby U20 Championship} in {\em Georgia}.
            \begin{itemize}
                \setlength{\parskip}{0cm}
                \setlength{\itemsep}{0cm}
                \item If the document provides enough information to conclude that there were several tries, but doesn't specify eight tries, that's fine.
            \end{itemize}
        \item {\em Shares} in the baby formula milk firm {\em Bellamy} have {\em plunged} after a warning that {\em new import regulations in China} will cut into revenues.
        \item {\em Iran's President Mahmoud Ahmadinejad has sacked Health Minister Marziyeh Vahid Dastjerdi}, the {\em sole woman in his cabinet}, state television reports.
        \end{itemize}

        \textbf{Summary should be reconstructable by the document's intended audience}
        \begin{itemize}
            \setlength{\parskip}{0cm}
            \setlength{\itemsep}{0cm}
            \item For many articles from the BBC news corpus, you may not have the contextual knowledge that the author assumes the audience to have. This is particularly glaring in the case of sports articles.
            \item We don't mean for you to have to Google proper nouns in order to do this annotation. If you can infer from the document that two expressions co-refer (e.g., ``Prime Minister'' in the summary and the individual's actual name in the document; country name in the summary and the specific town in the document), then you can consider the entity to be ``reconstructable'' even if you don't personally have the real-world knowledge to verify that the entities are the same.
            \item The exception is if you can't make sense of the article at all without doing a search. Please leave a comment on Anagram if you need to use a search engine to get relevant context in order to comprehend the basics of the article.
        \end{itemize}
        \\
    \bottomrule
    \end{tabular}
    \caption{Reconstructability judgment guideline.}
    \label{tab:reconstructability_guideline}
\end{table*}

\end{document}